\documentclass[11pt]{article}

\usepackage[final]{acl}

\usepackage{times}
\usepackage{latexsym}

\usepackage[T1]{fontenc}

\usepackage[utf8]{inputenc}

\usepackage{microtype}

\usepackage{inconsolata}

\usepackage{hyperref}
\usepackage{url}
\usepackage{amsmath}
\usepackage{graphicx}
\usepackage{float}
\usepackage{natbib}
\usepackage{algorithm}
\usepackage[noend]{algorithmic}
\usepackage{amsfonts}
\usepackage{color}
\usepackage{multirow}
\usepackage{booktabs}
\usepackage{makecell} 
\usepackage[table]{xcolor}
\usepackage{comment}
\usepackage{wrapfig}

\newcommand{\OUTPUT}{\item[\textbf{Output:}]}
\newcommand\blfootnote[1]{%
  \begingroup
  \renewcommand\thefootnote{}\footnote{#1}%
  \addtocounter{footnote}{-1}%
  \endgroup
}
\title{DREAM-S: Speculative Decoding with Searchable Drafting and Target-Aware Refinement for Multimodal Generation}

\author{%
  \textbf{Zining Liu$^{1,3}$\footnotemark[1] \quad Yunhai Hu$^{1}$ \quad Tianhua Xia$^{1}$ \quad Bo Bao$^{2}$} \quad  Eric Sather$^{2}$\\
  \textbf{Vithursan Thangarasa$^{2}$ \quad Sai Qian Zhang$^{1}$} \\
  $^1$New York University \\
  $^2$Cerebras Systems Inc. \quad $^3$University of Pennsylvania \\
  \texttt{zliu0@seas.upenn.edu,} \quad
  \texttt{\{yh5961, tx856, sai.zhang\}@nyu.edu,} \\
  \texttt{\{bo.bao, eric.sather, vithu\}@cerebras.net}
}

\begin{document}
\maketitle
\blfootnote{$^{*}$Work done during internship at New York University.}
\begin{abstract}
Speculative decoding (SD) has proven to be an effective technique for accelerating autoregressive generation in large language models (LLMs) however, its application to vision-language models (VLMs) remains relatively unexplored. We propose~\textit{DREAM-S}, a novel SD framework designed specifically for fast and efficient decoding in VLMs. DREAM-S leverages a neural architecture search (NAS) framework with target-aware supernet training to automatically identify both the optimal interaction strategy between the draft and target models, and the most suitable draft model architecture for the underlying hardware implementation platform. DREAM-S additionally incorporates adaptive intermediate feature distillation, guided by attention entropy, to enable efficient draft training. Experiments on a range of well-established VLMs show that DREAM-S achieves up to a $3.85\times$ speedup compared to standard decoding approaches and significantly outperforms existing SD baselines.
The code is publicly available at: \url{https://github.com/SAI-Lab-NYU/DREAM-S}.
\end{abstract}

\section{Introduction}
Vision-Language Models (VLMs) play a pivotal role in advancing artificial intelligence by integrating visual perception with natural language understanding. These models empower machines to process and generate both visual and textual data, enabling a broad array of applications such as image captioning~\citep{zhou2020unified, hu2022scaling, chen2022visualgpt, dzabraev2024vlrm}, visual question answering~\citep{chappuis2022prompt, bazi2023vision, wang2024surgical}, and content-based search~\citep{li2024searchlvlms, sun2025leveraging}. 
Despite their impressive capabilities, VLMs are computationally demanding, primarily due to the complexity of integrating high-dimensional visual and textual inputs. Speculative decoding (SD)~\citep{stern2018blockwise,leviathan2023fast} accelerates the autoregressive generation process of large language model (LLM) by dividing it into two stages: a low-cost drafting phase and a parallel verification phase. This allows multiple candidate tokens to be generated and then verified simultaneously in a single pass through the target LLM. The approach boosts decoding efficiency while keeping output quality via a selective acceptance-rejection mechanism.

As highlighted in prior work~\citep{chen2023accelerating, li2024eagle,li2024eagle2, cai2024medusa, ankner2024hydra, xia2023speculative, zhang2023draft, miao2023specinfer, chen2024sequoia,hu2025speculative}, achieving superior performance in the SD framework requires the draft model to meet two key criteria. First, it should achieve a high acceptance ratio, meaning that most of its proposed tokens are validated by the target model. Second, it should deliver low execution latency to minimize overall decoding time. Balancing these factors is essential for navigating the accuracy–latency trade-off, a challenge well-suited to neural architecture search (NAS), which has been widely demonstrated to yield highly effective trade-offs in similar settings. 

Although speculative decoding techniques have been widely developed to accelerate inference in LLMs, their integration into multimodal language models~\citep{li2024fast, raj2024faster}, especially VLMs, has received relatively little attention. In this paper, we introduce~\textit{Speculative Decoding with Searchable
Drafting and Target Aware Refinement for Multimodal Generation} (DREAM-S). Specifically, DREAM-S employs a neural architecture search (NAS) mechanism with target-guided distillation to train a supernet encompassing diverse draft configurations, then identify the optimal draft model configuration, input pruning ratio, and interaction strategy with the target model, all tailored to the underlying hardware platform. Another key innovation of DREAM-S is its selective utilization of intermediate-layer representations, which capture the most informative features from both modalities. These representations serve as effective supervision signals, enabling the draft model to achieve high accuracy. While neural architecture search and model pruning are well-established optimization techniques, their systematic application to draft model design in VLM speculative decoding remains unexplored. The key challenge lies not in developing new algorithms, but in formulating this multimodal acceleration problem within a principled search framework and designing VLM-specific optimization dimensions.
We evaluate DREAM-S on a diverse set of widely used vision-language models, including LLaVA-v1.6-Vicuna-7B/13B~\citep{liu2024improved}, SmolVLM-2B~\citep{marafioti2025smolvlm}, and Pixtral-12B~\citep{agrawal2024pixtral}, across a range of multimodal tasks. Extensive experiments show that DREAM-S significantly outperforms established speculative decoding baselines, while maintaining high acceptance rates across various applications. Our main contributions are as follows:

\begin{itemize} 
\item DREAM-S integrates a NAS framework to identify the optimal draft model configuration for optimal speedup. The search process further determines the optimal input and model pruning ratio and the most efficient connection strategy between the draft and target models for the optimal speedup performance. 
\begin{figure*}[!t]
  \begin{center}
    \includegraphics[width=1\textwidth]{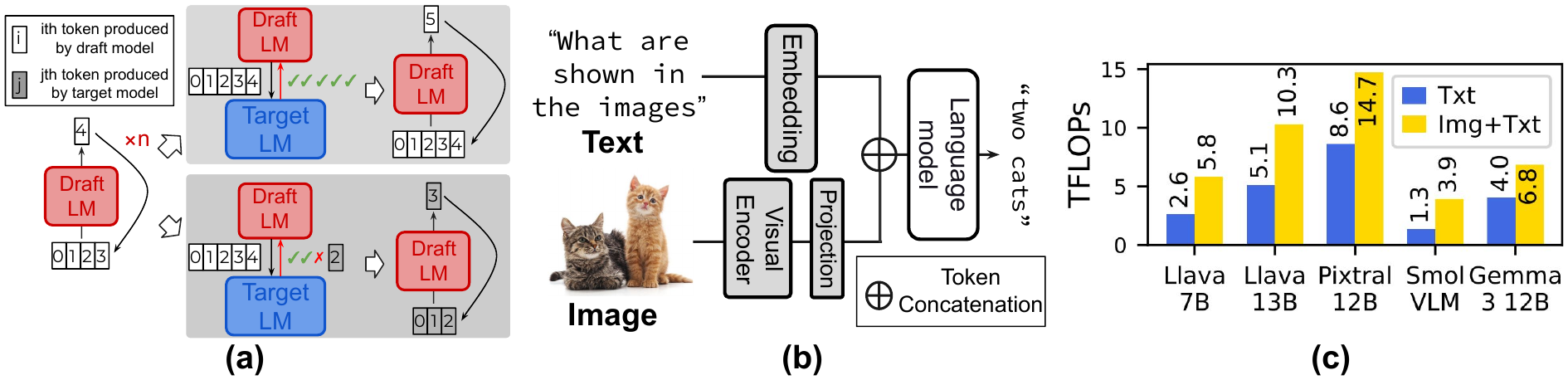}
  \end{center}
  \vspace{-10pt}
  \caption{(a) Speculative decoding process, LM denotes large model. (b) Architecture of VLM (c) Computational cost in Tera FLOPs of VLMs processing text only (Txt) and multi-modal (Img+Txt) inputs over different VLMs.}
  \label{fig:sd_intro}
\end{figure*}
\item During training, DREAM-S dynamically selects intermediate features from the target model’s middle layers based on proposed criteria, using them to supervise the draft model, which improves its predictive accuracy and extends token acceptance lengths. Additionally, DREAM-S employs a cross-attention mechanism to leverage these intermediate outputs, enabling more effective knowledge transfer from the target model and resulting in significant performance improvements.

\item Evaluation results demonstrate that DREAM-S is able to achieve up to a $3.85\times$ speedup compared to conventional decoding methods across various VLMs and tasks, surpassing existing speculative decoding approaches. 

\end{itemize}

\section{Related Work}
\label{sec:related-work}


\paragraph{Speculative decoding.} is an effective approach to alleviating the sequential bottleneck in language model inference~\citep{stern2018blockwise}. It divides the decoding process into two stages: a lightweight~\textit{draft model} quickly generates a sequence of candidate tokens, which are then verified in parallel by a more accurate~\textit{target model}, as illustrated in Figure~\ref{fig:sd_intro} (a). 

Let the draft model \( M_{da} \) generate \( \gamma \) draft tokens \( (t_1, \dots, t_\gamma) \) in each draft step. During the verification phase, the target model \( M_{ta} \) evaluates these tokens in parallel, but accepts them sequentially. If all tokens in the batch are accepted, the draft model proceeds to generate the next set of candidate tokens. (upper branch of Figure~\ref{fig:sd_intro}(a)). Otherwise, the target model supplies the correct token and assists the draft model in generating subsequent tokens (lower branch of Figure~\ref{fig:sd_intro}(a)). Specifically, it checks whether each draft token \( t_i \) matches the output of its own sampling. If a mismatch occurs at position \( i \), all tokens from \( t_i \) onward are discarded, and the target model's sampled token at position \( i \), denoted \( t'_i \), is used instead. The accepted token sequence is therefore \( (t_1, \dots, t_{i-1}, t'_i) \). 
SD allows for parallel token generation, moving beyond the conventional step-by-step decoding, while the verification phase ensures output quality by accepting or rejecting the draft tokens.


\paragraph{Vision-Language Models.} VLMs are designed to jointly process visual and textual inputs, enabling machines to interpret and generate content that integrates both modalities. As shown in Figure~\ref{fig:sd_intro}(b), a typical VLM consists of a~\textit{visual encoder} and a~\textit{language model}. The image is first processed by the visual encoder to extract visual tokens, which are then concatenated with textual tokens and passed into the language model to produce the final output. 
More recent models like LLaVA~\citep{liu2023visual}, InstructBLIP~\citep{dai2023instructblip}, and Pixtral~\citep{agrawal2024pixtral} focus on improving zero shot generalization by aligning model responses with human intent through instruction tuning.
While large VLMs achieve strong performance, their high computational cost and memory usage pose challenges for deployment on devices with limited resources. 

To quantify the computational cost introduced by visual input processing, we measure the FLOPs required by several models, including LLaVA-v1.6-Vicuna-7B, Pixtral-12B, and SmolVLM-2B, and Gemma3-12B~\citep{kamath2025gemma}, using the ScienceQA dataset. We select a representative example that includes a $480\times 300$ image, a prompt of 166 tokens, and a generated output of 500 tokens. FLOPs are computed using the PyTorch Profiler. As shown in Figure~\ref{fig:sd_intro}(c), processing both image and text inputs results in an average increase of $2.1~\times$ in computation compared to text-only inputs, highlighting the importance of developing more efficient visual processing methods. This additional cost does not only come from the one-time visual encoder prefill, but mainly from the autoregressive decoding stage: all visual tokens are stored in the KV cache and participate in attention at every decoding step, so each generated token must attend to both text and image tokens. 


\paragraph{Neural Architecture Search.}  By algorithmically exploring the vast architecture space, NAS alleviates the time-consuming process of training models with different configurations to find the most efficient and effective designs for specific tasks. Traditional NAS methods can be categorized into two classes. The first class of methods searches directly for the optimal architecture by making the search process differentiable. These approaches~\citep{liu2018darts,wan2020fbnetv2,chen2019progressive} formulate architecture search as a differentiable optimization problem. They apply continuous relaxation to express each operation as a weighted combination of candidate operations, allowing architecture parameters to be optimized jointly with network weights using gradient-based methods.
The second class of methods jointly trains a collection of nested neural networks and then employs a dedicated search network to select the optimal architecture from the trained candidates. Once For All (OFA)~\citep{cai2019once,chen2020dynamic,cai2018proxylessnas, zhang2022resnest} first trains a~\textit{SuperNet} containing diverse architectural configurations across four dimensions (depth, width, kernel size, and resolution), then applies progressive shrinking to train from large to small sub-networks. It uses a trained neural network and hardware-specific lookup tables to predict optimal sub-networks given target hardware and constraints such as latency budgets. 
DREAM-S employs target-aware supernet training, using intermediate features and distillation signals from the target model during draft training. This jointly optimizes computational efficiency and target alignment, achieving better speedup ratios.

\section{Methodology} 
\begin{figure*}[!t]
   \begin{center}
     \includegraphics[width=1\textwidth]{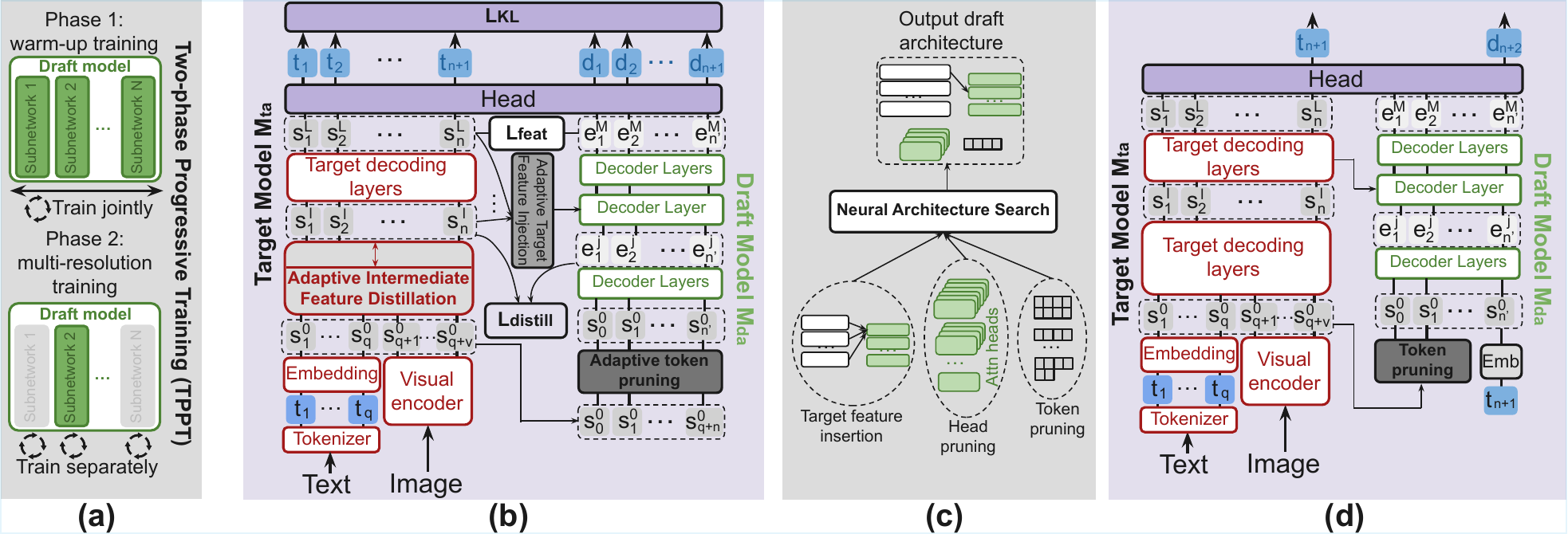}
   \end{center}
   \caption{DREAM-S framework overview. (a) Two-Phase Training: supernet training followed by subnetwork sampling. (b) Training with three losses: $\mathcal{L}_{KL}$, $\mathcal{L}_{feat}$, and $\mathcal{L}_{distill}$. (c) NAS exploring head pruning, token compression, and feature injection. (d) Inference operation of DREAM-S.}
   \label{fig:star-overall}
\end{figure*}

Figure~\ref{fig:star-overall} illustrates the training and inference workflows of DREAM-S. During training, the draft model processes multimodal inputs through two-phase progressive training (TPPT) and adaptive intermediate feature distillation (AIFD), detailed in Sections~\ref{sec:nas_architect} and~\ref{sec:intermedaite-distill}, respectively. During inference (Figure~\ref{fig:star-overall}d), DREAM-S selects the draft configuration that maximizes speedup on the target hardware. The framework leverages intermediate target features through cross-attention at inference time and knowledge distillation during training.

\subsection{Two-Phase Progressive Training}
\label{sec:nas_architect}
For the simplicity of interpretation, let $M_{ta}$ and $M_{da}$ denote the target and draft models, containing $L$ and $M$ transformer blocks, respectively. Let $t_n$ and $d_n$ be the $n$-th tokens generated by $M_{ta}$ and $M_{da}$. For the target model $M_{ta}$, the input consists of $q$ text prompt tokens and $v$ visual tokens. We define $s_n^{j-1}$ and $e_n^{j-1}$ as the intermediate hidden state of the $n$-th token at layer $j$ in $M_{ta}$ and $M_{da}$, respectively, with a total of $H_{j-1}$ attention heads.
Figure~\ref{fig:star-overall}(a) illustrates the Two-Phase Progressive Training (TPPT) procedure, which consists of two main stages. In the first, warm-up training phase, the entire draft model is trained. This includes the entire set of weights within $M_{da}$.


The training process is illustrated in detail in Figure~\ref{fig:star-overall}(b). At the initial decoding stage, the target model receives both textual and visual prompt tokens $t_1, \ldots, t_n$, where $n = q + v$, and begins predicting the subsequent token $t_{n+1}$. The draft model $M_{da}$ then predicts the $(n+2)$-th token, denoted as $d_{n+2}$. To enhance the quality of draft token generation, we integrate intermediate features from both models using a cross-attention mechanism. Specifically, we extract features from a selected layer $l$ of the target model $M_{ta}$, represented as $S^{l} = (s^{l}_1, s^{l}_2, \dots, s^{l}_n)$, and from the $j$-th layer of the draft model $M_{da}$, shown as $E^j = (e^j_1, e^j_2, \dots, e^j_{n'})$, where $n' = q + r$, and $0 \leq r \leq v$ is the number of visual tokens fed to the draft model. In this cross-attention setup, $E^j$ serves as query, $S^{l}$ provides the keys and values.

The second phase,~\textit{multi-resolution training}, utilizes the OFA framework to train draft subnetworks. Our NAS framework operates across three dimensions grounded in established transformer optimization principles, and we validate the necessity of searching each dimension through preliminary experiments on LLaVA-v1.6-Vicuna-7B. First, attention head pruning leverages observations that transformer attention heads exhibit varying importance for model performance~\citep{michel2019sixteen, voita2019analyzing, xia2023sheared}. We validate our search dimensions by pruning 25\% of attention heads and 30\% of visual tokens, each yielding a 6\% speedup, as shown in Table~\ref{tab:compression-ablation}. In addition, DREAM-S explores adaptive target feature injection, searching for optimal extraction layers rather than relying on fixed positions as in prior methods such as EAGLE~\citep{li2024eagle, li2024eagle2}.

\begin{table}[!t]
  \centering
  \small
  \resizebox{0.4\textwidth}{!}{
    \begin{tabular}{l|cc}
\toprule
\multirow{2}{*}{Configuration} & \multicolumn{2}{c}{MMT-Bench} \\
& S & $\tau$  \\
\midrule
Vanilla & 2.45 & 6.50  \\
Head pruning (0.75) & 2.57 & 6.31  \\
Input compression (0.7) & 2.56 & 6.40 \\
\bottomrule
\end{tabular}
}
\caption{Impact of head pruning and input compression on speculative decoding performance.}
\label{tab:compression-ablation}
\end{table}

In addition, we explore the choice of layer in the target model $M_{ta}$ from which features are extracted for the cross-attention mechanism in the draft model $M_{da}$, as illustrated in the \textit{Adaptive Target Feature Injection} module in Figure~\ref{fig:star-overall}(b). The OFA search process is illustrated in Figure~\ref{fig:star-overall}(c), where during each training iteration of DREAM-S, a draft subnetwork with a specific visual token budget $r*$, attention head configuration $\textbf{H} = \{H_j\}, 1\leq j\leq M$, and selected connection layer $l$ is randomly sampled. This draft model, denoted as $M_{da}(r, \textbf{H}, l)$, is then trained following the procedure described in Figure~\ref{fig:star-overall}(b).

\paragraph{Attention Head-Wise Pruning.}
\label{sec:head_prune}

DREAM-S dynamically computes head importance without storing persistent rankings. Let $H'_{j}$ (where $1 \leq H'_{j} \leq H_{j}$) represent the number of attention heads retained at the $j$-th layer. During training, for each subnetwork configuration, we select and retain a top $H'_{j}$ attention heads with the highest importance scores.
To evaluate the importance $I_{j}$ of each attention head $h$, we aggregate the product of gradients and their corresponding weight magnitudes across all projection matrices associated with that head~\citep{michel2019sixteen, molchanov2016pruning}:
\begin{equation}
\label{eqn:attention_gradient_weight}
    \small
    I_{j} = \sum_{P \in \{Q,K,V\}} \sum_{x,y}  \left| \nabla W_{P}^{h,j} [x,y] \cdot W_{P}^{h,j} [x,y] \right|
\end{equation}
Here, $W_{P}^{h,j}$ denotes projection matrices (query (Q), key (K), or value (V)) associated with head $h$ at layer $j$, and $\nabla W_{P}^{h,j}$ represents its corresponding gradient. The notation $[x, y]$ refers to the element located at the $x$-th row and $y$-th column of the matrix.
In each training iteration, attention heads are ranked using Equation~\ref{eqn:attention_gradient_weight}, and a subset is selected according to the specified budget $H_{j}$.

\paragraph{Visual Token Compression.}
\label{sec:visual_comp}
DREAM-S adopts a target-aware token selection mechanism that leverages the attention patterns of $M_{ta}$ to guide this visual token selection. DREAM-S evaluates the visual token importance using attention scores from the target model's final layer during the prefilling phase: Given target model attention weights $A^{(L)} \in \mathbb{R}^{B \times H \times Q \times K}$ from the final layer, the importance score for visual token $j$ is:
\begin{equation}
    \small
    I_j = \frac{1}{H \cdot Q} \sum_{h=1}^{H} \sum_{i=1}^{Q} A^{(L)}_{h,i,j}
\end{equation}
In each forward pass, a token budget $r$ is randomly sampled from the budget pool $\mathcal{R}$. Based on this budget, only $r$ visual tokens are retained from the input according to their importance scores, while the remaining tokens remain unchanged. This selective compression preserves informative visual features while reducing computational cost.

\paragraph{Adaptive Target Feature Injection.}

DREAM-S explores the performance impact of selecting different target layers. We consistently use final-layer target features for supervision during warmup training, then systematically explore different target layer choices during multi-resolution training to optimize draft performance within our NAS framework (Figure~\ref{fig:star-overall}(b)).
In each training iteration, a target feature is randomly sampled from one of the candidate layers in the target model and injected into the draft model. To incorporate this feature, the draft model includes a cross-attention layer inserted at a predefined layer index. Specifically, we integrate the target features from a selected layer $l'$ of $M_{ta}$, denoted as $S^{l'} = (s^{l'}_1, s^{l'}_2, \dots, s^{l'}_n)$, with the intermediate features of the $j$-th layer of the draft model $M_{da}$, denoted as $E^{j} = (e^j_1, e^j_2, \dots, e^j_{n'})$, via a cross-attention mechanism. $E^{j}$ serves as the query, while $S^{l'}$ acts as the keys and values.

\subsection{Adaptive Intermediate Feature Distillation}
\label{sec:intermedaite-distill}
Beyond the NAS framework detailed in Section~\ref{sec:nas_architect}, DREAM-S adaptively selects intermediate features from the target model $M_{ta}$ for distillation into the draft model's early layers using a dedicated loss function. To effectively guide the training of the draft model $M_{da}$, the selected target features must meet two essential criteria. They should capture semantically meaningful content and exhibit low variability across layers to ensure stable learning. Prior studies~\citep{sun2020contrastive, skean2025layer, jain2024elevating} show that intermediate features with low attention entropy and high consistency provide stable supervision signals. DREAM-S adopts a simple yet effective strategy to identify such features from each layer of the target model, as illustrated in Figure~\ref{fig:star-overall}(b), to support the efficient training of the draft model. However, low entropy alone is insufficient since a layer may exhibit low entropy while fluctuating strongly between adjacent layers, causing unstable supervision. DREAM-S therefore jointly considers both the entropy value ($\text{AE}(\ell)$) and its inter-layer variation $\Delta \text{AE}(\ell)$, selecting layers that are information rich and consistent across depth for robust knowledge transfer.

Specifically, for the $l$-th decoder block of $M_{ta}$, its input tokens and output tokens are denoted as $S^{\ell-1} = \bigl(s_1^{\ell-1}, s_2^{\ell-1}, \dots, s_n^{\ell-1}\bigr)$ and $S^{\ell} = \bigl(s_1^{\ell}, s_2^{\ell}, \dots, s_n^{\ell}\bigr)$, respectively. Let the attention matrix $A_{\ell}$ associated with the $l$-th layer be $A_{\ell} = \operatorname{softmax}(\frac{Q_{\ell} K_{\ell}^\top}{\sqrt{z}})$, where $Q_{\ell} = S^{\ell-1} W_Q$ and $K_{\ell} = S^{\ell-1} W_K$, and $z$ denotes the hidden dimension of the $M_{da}$. The average attention entropy (AE) is calculated as 
$\small\text{AE}(\ell)=
  -\frac{1}{n}\sum_{i=1}^{n}\sum_{j=1}^{n}
      A_{\ell, i,j}
      \log A_{\ell, i,j}$
where $A_{\ell, i,j}$ denotes the $(i,j)$-th element of $A_{\ell}$. In practice with multiple heads, $\text{AE}(\ell)$ is also averaged across the selected attention heads of the subnetwork. To capture variation across layers, we further define the one-step difference \(\Delta \text{AE}(\ell) = \bigl| \text{AE}(\ell) - \text{AE}(\ell-1) \bigr|.\) By jointly considering both $\text{AE}(\ell)$ and its inter-layer variation $\Delta \text{AE}(\ell)$, we identify the optimal distillation layer \( \ell^{*}_{d} = \operatorname*{arg\,min}_{\ell \in L}\bigl[\Delta \text{AE}(\ell) + \text{AE}(\ell)\bigr],\) ensuring transferred features from $M_{ta}$ to $M_{da}$ are semantically rich and locally stable.

\begin{table*}[!t]
    \vspace{0pt}
    \centering\resizebox{1.8\columnwidth}{!}{
    {\huge
       \begin{tabular}{cccccccccccccc|cc}
        \toprule
        &  & \multicolumn{2}{c}{MMT} & \multicolumn{2}{c}{SEED} & \multicolumn{2}{c}{ScienceQA} & \multicolumn{2}{c}{OCRBench} & \multicolumn{2}{c}{ChartQA} & \multicolumn{2}{c}{MathVista}  & \multicolumn{2}{|c}{Average} \\   \midrule
    Models & Methods & S & $\tau$ & S & $\tau$ & S & $\tau$ & S & $\tau$ & S & $\tau$ & S & $\tau$ & S & $\tau$   \\ \midrule
    \multicolumn{16}{c}{Temperature = 0} \\ \midrule
    \multirow{8}{*}{\rotatebox{0}{\makecell{LLaVA-v1.6 \\Vicuna-7B}}} & SPD~\citep{gagrani2024speculative} &  1.10 & 1.88 & 0.81 & 1.17 & 1.08 & 1.87 & 0.89 & 1.25 & 0.91 & 1.24  & 1.06 & 1.76     & 0.97 & 1.53\\
     & Kangaroo~\citep{liu2024kangaroo} &  1.32 & 2.11 & 1.33 & 2.12 & 1.31 & 2.09 & 1.17 & 1.89 & 1.18 & 1.98  & 1.15 & 1.86    & 1.24 & 2.01\\
     & Medusa~\citep{cai2024medusa} & 1.58 & 2.88 & 1.59 & 3.01 & 1.44 & 2.77 & 1.22 & 2.33 & 1.25 & 2.41  & 1.22 & 2.34     & 1.38 & 2.62 \\
     & Hydra~\citep{ankner2024hydra} & 1.78 & 3.86 & 1.72 & 3.88 & 1.68 & 3.79 & 1.41 & 3.21 & 1.35 & 3.11  & 1.42 & 3.25    & 1.56 & 3.52 \\
     & EAGLE~\citep{li2024eagle} & 2.10 & 5.04 & 2.09 & 5.01 & 1.98 & 4.88 & 1.72 & 4.13 & 1.56 & 3.98 & 1.78 & 4.25  & 1.87 & 4.55 \\
     & EAGLE-2~\citep{li2024eagle2} & 2.31 & 5.48 & 2.31 & 5.61 & 2.15 & 5.22 & 1.92 & 4.88 & 1.77 & 4.22  & 1.87 & 4.67   & 2.05 & 5.01 \\
    
     & EAGLE-3~\citep{li2025eagle} & 2.38 & 5.72 & 2.36 & 5.82 & 2.22 & 5.52 & 2.02 & 5.24 & 1.83 & 4.46 & 1.97 & 5.02 & 2.13 & 5.30 \\
       & DREAM ~\citep{hu2025dream} & 2.52 & 6.40 & 2.48 & 6.20 & 2.33 & 5.82 & 2.05 & 4.88 & 1.89 & 4.44  & 2.11 & 5.32  & 2.23 & 5.51 \\
     \rowcolor{blue!10}
     \cellcolor{white!10} & \textbf{DREAM-S}  & \textbf{2.67} & 6.27  & \textbf{2.61} & 6.18  & \textbf{2.45} & 5.71  & \textbf{2.11} & 4.89  & \textbf{2.04} & 4.39  & \textbf{2.20} & 5.30   & \textbf{2.35} & 5.46\\
\midrule

     \multirow{8}{*}{\rotatebox{0}{\makecell{LLaVA-v1.6 \\Vicuna-13B}}} & SPD & 1.07 & 1.78 & 1.06 & 1.79 & 1.09 & 1.88 & 0.86 & 1.12 & 0.89 & 1.25  & 0.87 & 1.22     & 1.00 & 1.58\\
    & Kangaroo &  1.43 & 1.77 & 1.51 & 1.87 & 1.22 & 1.55 & 1.21 & 1.54 & 1.27 & 1.61 & 1.53 & 2.01    & 1.36 & 1.72\\
     & Medusa & 1.99 & 2.67 & 1.96 & 2.76 & 1.93 & 2.77 & 1.40 & 2.92 & 1.51 & 2.82 & 1.51 & 2.62    & 1.72 & 2.76  \\
     & Hydra & 2.12 & 2.87 & 2.08 & 2.99 & 2.21 & 3.12 & 1.49 & 3.07 & 1.65 & 3.03 & 1.66 & 2.87   & 1.87 & 2.99 \\
     & EAGLE &  2.45 & 3.56 & 2.19 & 3.24 & 2.63 & 3.98 & 1.65 & 3.31 & 1.85 & 3.27 & 1.8 & 3.09   & 2.10 & 3.41 \\
     & EAGLE-2 & 2.89 & 4.05 & 3.18 & 4.33 & 3.09 & 4.97 & 2.20 & 4.12 & 2.41 & 4.15  & 2.39 & 3.76   & 2.69 & 4.23 \\
     & EAGLE-3 & 3.45 & 4.90 & 3.34 & 4.65 & 3.19 & 5.20 & 2.50 & 4.79 & 2.46 & 4.37 & 2.42 & 3.85 & 2.89 & 4.63 \\
     & DREAM & 3.68 & 5.58  & 3.51 & 5.34  & 3.36 & 5.29  & 2.69 & 4.64  & 2.59 & 4.20  & 2.53 & 4.18    & 3.06 & 4.87\\
    \rowcolor{blue!10}
    \cellcolor{white!10}& \textbf{DREAM-S} & \textbf{3.85} & 5.56  & 
    \textbf{3.61} & 5.32  & 
    \textbf{3.41} & 5.19  & 
    \textbf{2.77} & 4.61  & 
    \textbf{2.67} & 4.17  & 
    \textbf{2.62} & 4.11  & 
    \textbf{3.16} & 4.82\\
    \midrule

    \multirow{8}{*}{\rotatebox{0}{\makecell{Pixtral-12B}}} & SPD  & 1.08 & 1.51  & 1.03 & 1.47  & 1.05 & 1.49  & 1.05 & 1.49  & 1.04 & 1.43  & 1.04 & 1.46   & 1.05 & 1.47\\
    & Kangaroo  & 1.26 & 1.54  & 1.09 & 1.39  & 1.14 & 1.51  & 1.16 & 1.52  & 1.12 & 1.47   & 1.13 & 1.49   & 1.15 & 1.49\\
    & Medusa  & 1.37 & 1.81  & 1.37 & 1.81  & 1.35 & 1.87  & 1.24 & 1.69  & 1.22 & 1.68    & 1.16 & 1.47   & 1.28 & 1.72\\
    & Hydra  & 1.58 & 2.24  & 1.47 & 2.04  & 1.53 & 2.06  & 1.38 & 1.81  & 1.34 & 1.79    & 1.36 & 1.78   & 1.44 & 1.95\\
    & EAGLE  & 2.38 & 3.47  & 1.97 & 2.53  & 2.31 & 3.64  & 1.69 & 2.73  & 1.78 & 2.84   & 1.64 & 2.47   & 1.96 & 2.95\\
    & EAGLE-2  & 2.81 & 3.95  & 2.31 & 3.07  & 2.64 & 4.03  & 2.12 & 3.25  & 2.14 & 3.17   & 1.81 & 2.73    & 2.31 & 3.37\\
    & EAGLE-3 & 2.83 & 4.12 & 2.46 & 3.40 & 2.79 & 4.41 & 2.22 & 3.48 & 2.26 & 3.51 & 2.13 & 3.38 & 2.45 & 3.72 \\
    & DREAM  & 2.93 & 4.52  & 2.61 & 3.67  & 2.98 & 4.33  & 2.38 & 3.55   & 2.35 & 3.49    & 2.36 & 3.42    & 2.60 & 3.78\\
    \rowcolor{blue!10}
    \cellcolor{white!10}& \textbf{DREAM-S}  & \textbf{3.01} & 4.41  & \textbf{2.73} & 3.56  & \textbf{3.09} & 3.93  & \textbf{2.46} & 3.44   & \textbf{2.40} & 3.42    & \textbf{2.42} & 3.34    & \textbf{2.69} & 3.68\\
    \midrule

    \multirow{8}{*}{SmolVLM-2B} & SPD & 1.02 & 1.33  & 1.04 & 1.41  & 1.06 & 1.43  & 1.06 & 1.42  & 1.07 & 1.46  & 1.02 & 1.34  & 1.04 & 1.40\\
    & Kangaroo  & 1.28 & 1.48  & 1.08 & 1.18  & 1.03 & 1.17  & 1.06 & 1.22  & 1.04 & 1.14   & 1.08 & 1.23   & 1.10 & 1.24\\
    & Medusa  & 2.12 & 2.71  & 1.51 & 2.00  & 1.72 & 2.22  & 1.20 & 1.61  & 1.15 & 1.55   & 1.35 & 1.75    & 1.51 & 1.97\\
    & Hydra  & 2.33 & 3.07  & 1.62 & 2.08  & 1.98 & 2.66  & 1.32 & 1.74  & 1.22 & 1.58   & 1.51 & 1.98    & 1.66 & 2.19\\
    & EAGLE  & 2.57 & 3.42  & 1.85 & 2.56  & 2.16 & 2.76  & 1.42 & 1.88  & 1.34 & 1.77    & 1.65 & 2.22    & 1.83 & 2.44\\
    & EAGLE-2  & 2.96 & 3.89  & 2.12 & 2.93  & 2.39 & 3.21  & 1.65 & 2.11  & 1.51 & 2.13   & 1.81 & 2.63    & 2.07 & 2.82\\
    & EAGLE-3 & 3.00 & 3.94 & 2.17 & 3.04 & 2.65 & 3.57 & 1.78 & 2.33 & 1.60 & 2.30 & 1.97 & 2.84 & 2.20 & 3.00 \\
    & DREAM  & 3.05 & 3.97  & 2.24 & 3.18  & 2.85 & 3.62 & 1.85 & 2.56 & 1.62 & 2.33 & 2.01 & 2.88  & 2.27 & 3.09\\
    \rowcolor{blue!10}
    \cellcolor{white!10}& \textbf{DREAM-S}  & \textbf{3.12} & 3.94  & \textbf{2.28} & 3.16  & \textbf{2.91} & 3.57  & \textbf{1.88} & 2.51  & \textbf{1.64} & 2.28   & \textbf{2.06} & 2.82     & \textbf{2.32} & 3.05\\ 
    \midrule
    \multicolumn{16}{c}{Temperature = 1} \\ 
    \midrule

    \multirow{4}{*}{\rotatebox{0}{\makecell{LLaVA-v1.6 \\Vicuna-7B}}} & SPD  & 0.83 & 1.19  & 0.81 & 1.15  & 0.85 & 1.18  & 0.75 & 1.06  & 0.72 & 1.08   & 0.92 & 1.48    & 0.81 & 1.19\\
    & Kangaroo  & 1.20 & 1.97  & 1.26 & 2.03  & 1.23 & 2.01  & 1.09 & 1.80  & 1.11 & 1.89   & 1.07 & 1.77    & 1.16 & 1.91\\
    & EAGLE-2  & 2.19 & 5.37  & 2.20 & 5.48  & 2.04 & 5.12  & 1.82 & 4.77  & 1.68 & 4.13  & 1.76 & 4.56   & 1.95 & 4.91\\
    & EAGLE-3 & 2.25 & 5.70 & 2.25 & 5.72 & 2.10 & 5.38 & 1.89 & 5.01 & 1.71 & 4.28 & 1.88 & 4.98 & 2.01 & 5.18 \\
    & DREAM & 2.39 & 6.29  & 2.35 & 6.07  & 2.25 & 5.68  & 1.99 & 4.88  & 1.84 & 4.41  & 2.02 & 5.23  & 2.14 & 5.43\\
    \rowcolor{blue!10}
    \cellcolor{white!10}& \textbf{DREAM-S}   & \textbf{2.50} & 6.25  & 
    \textbf{2.45} & 5.92  & 
    \textbf{2.33} & 5.56  & 
    \textbf{2.03} & 4.75  & 
    \textbf{1.97} & 4.22  & 
    \textbf{2.09} & 5.13    & \textbf{2.23} & 5.30\\
    \midrule

     \multirow{4}{*}{\rotatebox{0}{\makecell{LLAVA-v1.6 \\Vicuna-13B}}} & SPD &  0.88 & 1.22 & 0.84 & 1.25 & 0.84 & 1.32 & 0.79 & 1.18 & 0.81 & 1.14 & 0.88 & 1.24   & 0.84 & 1.22\\
          & Kangaroo  & 1.23 & 1.57  & 1.17 & 1.53  & 1.07 & 1.44  & 1.01 & 1.24  & 1.07 & 1.34  & 1.21 & 1.67    & 1.13 & 1.46\\
    & EAGLE-2  & 2.35 & 3.75  & 3.02 & 4.30  & 3.03 & 4.67  & 2.03 & 3.87  & 2.18 & 3.83  & 2.18 & 3.41    & 2.46 & 3.97\\
    & EAGLE-3 & 2.92 & 4.77 & 3.12 & 4.61 & 3.06 & 4.89 & 2.08 & 4.03 & 2.26 & 4.04 & 2.19 & 3.55 & 2.61 & 4.32 \\
    & DREAM  & 3.34 & 5.38  & 3.32 & 5.06  & 3.20 & 5.98  & 2.22 & 3.89  & 2.43 & 4.04  & 2.29 & 4.03 & 2.80 & 4.73\\
    \rowcolor{blue!10}
    \cellcolor{white!10}& \textbf{DREAM-S}  & 
    \textbf{3.51} & 5.37  & 
    \textbf{3.55} & 5.00  & 
    \textbf{3.38} & 5.88  & 
    \textbf{2.35} & 3.92 & 
    \textbf{2.59} & 4.09  & 
    \textbf{2.38} & 3.99  & 
    \textbf{2.96} & 4.71\\
    \midrule

     \multirow{4}{*}{\rotatebox{0}{\makecell{Pixtral-12B}}} & SPD  & 0.81 & 1.15  & 0.79 & 1.11  & 0.80 & 1.12  & 0.80 & 1.13  & 0.75 & 1.07  & 0.77 & 1.09    & 0.79 & 1.11\\
    & Kangaroo  & 1.18 & 1.41  & 1.08 & 1.35  & 1.03 & 1.36  & 1.19 & 1.48  & 1.14 & 1.45  & 1.09 & 1.41    & 1.12 & 1.41\\
    & EAGLE-2  & 2.76 & 3.81  & 2.24 & 3.01  & 2.76 & 3.87  & 2.23 & 3.24  & 2.03 & 3.09  & 1.79 & 2.69    & 2.30 & 3.28\\
    & EAGLE-3 & 2.79 & 4.02 & 2.33 & 3.25 & 2.80 & 4.03 & 2.25 & 3.51 & 2.27 & 3.58 & 1.92 & 2.98 & 2.39 & 3.56 \\
    & DREAM & 2.90 & 4.02  & 2.47 & 3.57  & 2.93 & 3.94  & 2.29 & 3.46  & 2.21 & 3.21   & 2.16 & 3.27 & 2.49 & 3.58\\
    \rowcolor{blue!10}
    \cellcolor{white!10}& \textbf{DREAM-S} & \textbf{2.98} & 3.93  & \textbf{2.56} & 3.48  & \textbf{2.99} & 3.79  & \textbf{2.34} & 3.32  & \textbf{2.26} & 3.09   & \textbf{2.22} & 3.22      & \textbf{2.56} & 3.47\\
    \midrule

     \multirow{4}{*}{SmolVLM-2B} & SPD & 1.07 & 1.47  & 1.01 & 1.33  & 1.07 & 1.46  & 0.97 & 1.26  & 1.06 & 1.44   & 0.85 & 1.20     & 1.00 & 1.36\\ 
     & Kangaroo  & 1.37 & 1.59  & 1.12 & 1.24  & 1.22 & 1.41  & 1.12 & 1.29  & 1.18 & 1.36  & 1.28 & 1.42    & 1.22 & 1.39\\
    & EAGLE-2  & 2.62 & 3.60  & 1.92 & 2.67  & 2.24 & 3.11  & 1.41 & 1.77  & 1.60 & 2.18  & 1.77 & 2.49    & 1.93 & 2.64\\
    & EAGLE-3 & 2.77 & 3.82 & 2.11 & 3.04 & 2.63 & 3.65 & 1.46 & 1.90 & 1.64 & 2.29 & 1.84 & 2.64 & 2.08 & 2.89 \\
    & DREAM & 2.88 & 3.66  & 2.25 & 3.33  & 2.91 & 3.74  & 1.54 & 2.12  & 1.77 & 2.51& 1.97 &  2.70 & 2.22 & 3.01  \\
    \rowcolor{blue!10}
    \cellcolor{white!10}& \textbf{DREAM-S}  & \textbf{2.93} & 3.61  & \textbf{2.33} & 3.30  & \textbf{2.96} & 3.67  & \textbf{1.59} & 2.12  & \textbf{1.81} & 2.48& \textbf{2.01} &  2.66   & \textbf{2.27} & 2.97   \\
    \bottomrule
    \end{tabular}}
}
\vspace{2pt} 
\caption{Evaluation of DREAM-S on speedup ratio (S) and average accepted token length ($\tau$).}
\label{tab:eval:main}
\end{table*}

\subsection{TPPT Loss Design}
\label{sec:tppt_summary}
This section presents the design of the TPPT loss function. We employ a multi-component weighted loss function to align draft model with target model across multiple levels of representation, where $\lambda$ terms control relative importance of each component. The loss function comprises three terms: (1) a KL divergence loss $\mathcal{L}_{\mathrm{KL}} = \mathrm{KL}(\mathrm{softmax}(D), \mathrm{softmax}(T))$ that ensures output token distributions match between the draft and target models, where $D = (d_1, \dots, d_n)$ and $T = (t_1, \dots, t_n)$ represent the predicted token logits from the draft model $M_{da}$ and target model $M_{ta}$; (2) an intermediate feature distillation loss $\mathcal{L}_{distill} = \mathrm{smoothL1}(E^m, S^{\ell^\star})$ that aligns early-layer features from the draft model ($E^m$ with $m=1$) with selected intermediate features from the target model ($S^{\ell^{*}_{d}}$); and (3) a feature alignment loss $\mathcal{L}_{feat} = \mathrm{smoothL1}(E^{M}, S^{L})$ that matches the final-layer features between the draft model's output $E^{M} = (e_1^{M}, \dots, e_n'^{M})$ and the target model's output $S^{L} = (s_1^{L}, \dots, s_n^{L})$ to improve token acceptance rates. The loss $\mathcal{L}_{final}$ for TPPT is: 
\begin{equation}
    \small
    \mathcal{L}_{final} = \lambda_{feat}\,\mathcal{L}_{feat} + \lambda_{distill}\,\mathcal{L}_{distill} + \lambda_{KL}\,\mathcal{L}_{\mathrm{KL}}
    \label{eq:loss}
\end{equation} 

\renewcommand{\arraystretch}{1.15} 

\section{Results}

\paragraph{Experimental Setup.}  We assess DREAM-S across four VLMs spanning different parameter scales: LLaVA-v1.6-Vicuna (7B, 13B)~\citep{liu2024llavanext}, Pixtral (12B)~\citep{agrawal2024pixtral} and SmolVLM (2B)~\citep{marafioti2025smolvlm}. Evaluation is conducted on six multimodal benchmarks: MMT-Bench~\citep{ying2024mmt}, SEED-Bench-2~\citep{li2023seed2}, ScienceQA~\citep{lu2022learn}, OCRBench~\citep{liu2024ocrbench}, ChartQA~\citep{masry2022chartqa}, and MathVista~\citep{lu2024mathvista}.
We measure two metrics: (1) \textbf{Speedup ratio} calculated as $t_{\mathrm{AR}} / t_{\mathrm{method}}$, where $t_{\mathrm{AR}}$ represents the average wall-clock time per token for standard autoregressive decoding and $t_{\mathrm{method}}$ denotes the time for each evaluated approach. 
(2) \textbf{Average token acceptance length} $\tau$,  quantifying consecutive draft tokens accepted during verification. 
We implement six state-of-the-art SD methods adapted for VLMs: SPD~\citep{gagrani2024speculative}, Kangaroo~\citep{liu2024kangaroo}, Medusa~\citep{cai2024medusa}, Hydra~\citep{ankner2024hydra}, EAGLE 1, 2, 3~\citep{li2024eagle, li2024eagle2, li2025eagle} and Dream~\citep{hu2025dream}.
Target VLMs remain frozen while draft models undergo training. We utilize the LLaVA-\texttt{mix665k} dataset with 55,000 training samples, supplemented by 1,000 samples from each evaluation benchmark that are~\textbf{disjoint} from test sets for domain adaptation. 

For multi-resolution training in TPPT, the visual token pruning budget pool is defined as $\mathcal{R} = \{0.1n, 0.2n, \ldots, n\}$, where $n$ denotes the total number of prompt tokens. The head pruning configuration is set as $H_{j} = \{0.25H, 0.5H, 0.75H, H\}$, where $H_{j}$ is the number of retained attention heads and $H$ is the total number of heads. Adaptive token feature injection searches across the last five layers of the target model to determine the optimal layer index for injecting features into the draft model. After TPPT is complete, the optimal draft model is selected by exhaustively searching through all subnetwork candidates to identify the one that achieves the highest speedup.
The draft architecture comprises three decoder layers. Loss term weights are configured as: $\mathcal{L}_{\mathrm{feat}} = 0.2$, $\mathcal{L}_{\mathrm{distill}} = 0.2$, and $\mathcal{L}_{\mathrm{KL}} = 1.0$. All experiments run on a single NVIDIA A100 80GB GPU. 

Both phases of TPPT are trained for 68,000 iterations using AdamW ($\beta_1 = 0.9$, $\beta_2 = 0.95$) with a learning rate of $3 \times 10^{-5}$ and gradient clipping at 0.5. All experiments follow standard speculative decoding practice with inference batch size 1, reporting wall-clock time per generated token over the full decoding pipeline. All methods are implemented in PyTorch 2.0.1 with HuggingFace Transformers 4.36.2 under CUDA 12.4.

Regarding training cost, Phase 1 supernet training requires approximately 3.5 hours per epoch on four NVIDIA A100 80GB GPUs for LLaVA-v1.6-Vicuna-7B, which is identical to the cost of standard single-draft fine-tuning as in EAGLE. Phase 2 subnet training adds approximately 1 additional hour per epoch due to dynamic pruning operations. The NAS search is conducted offline once per model-hardware pair, taking roughly 12 minutes per 100 mini-batches on a single A100, and is not repeated during deployment. This one-time cost is fully amortized over all subsequent inference on that hardware platform and remains negligible compared to standard VLM pretraining.
\subsection{Main Results and Discussion}
Table~\ref{tab:eval:main} demonstrates DREAM-S's performance across four VLMs and six multimodal benchmarks, showing consistent superiority over existing speculative decoding methods in both speedup ratios S and token acceptance lengths $\tau$. DREAM-S achieves substantial acceleration across all evaluated models, with average speedups ranging from $2.32\times$ to  $3.16\times$ compared to standard autoregressive decoding. 
 Most notably, DREAM-S consistently outperforms the strongest existing baselines, including both EAGLE-3 and DREAM. Compared to EAGLE-3, DREAM-S achieves higher average speedups across all evaluated models, with improvements of 10\% on LLaVA-7B ($2.35\times$ vs $2.13\times$), 9\% on LLaVA-13B ($3.16\times$ vs $2.89\times$), 10\% on Pixtral-12B ($2.69\times$ vs $2.45\times$), and 5\% on SmolVLM-2B ($2.32\times$ vs $2.20\times$). DREAM-S also consistently surpasses DREAM, yielding additional speedups of 2–5\% across all models, including LLaVA-7B ($2.35\times$ vs.\ $2.23\times$), LLaVA-13B ($3.16\times$ vs.\ $3.06\times$), and Pixtral-12B ($2.69\times$ vs.\ $2.65\times$).
Interestingly, while DREAM-S's token acceptance lengths are sometimes comparable to or slightly lower than DREAM (e.g., 6.27 vs 6.40 on LLaVA-7B for MMT-Bench, 5.19 vs 5.29 on LLaVA-13B for ScienceQA), DREAM-S achieves higher speedups through attention head pruning and visual token compression, which reduce draft model FLOPs. This creates a favorable trade-off where slightly lower or comparable draft quality is more than compensated by substantially more efficient token generation.

Larger models demonstrate greater benefits from DREAM-S's optimizations. The 13B LLaVA model achieves the highest speedup of $3.16\times$, compared to $2.35\times$ for the 7B variant. This aligns with the expectation that computationally heavier target models create more opportunities for draft model acceleration. Pixtral-12B shows competitive performance despite its larger parameter count, suggesting that DREAM-S's target-aware compression effectively handles diverse architectural designs.

DREAM-S exhibits varying effectiveness across different benchmark types. Vision-language reasoning tasks like MMT-Bench and ScienceQA yield the highest speedups (e.g., $2.67\times$ and $2.45\times$ respectively on LLaVA-7B), as these tasks benefit from DREAM-S's ability to capture semantic relationships between visual and textual content. Conversely, tasks requiring precise visual detail recognition, such as OCRBench, show modest improvements, reflecting the difficulty of compressing visual information without accuracy loss.

Temperature settings critically impact performance, with DREAM-S achieving optimal results at deterministic decoding $T = 0$ but degrading moderately under stochastic sampling $T = 1$ due to increased token variance. Nevertheless, on LLaVA-v1.6-Vicuna-7B, DREAM-S maintains superior performance with $2.23\times$ speedup and $\tau = 5.30$ versus EAGLE-3's $2.01\times$ speedup and  $\tau = 5.18$ and Dream's $2.14\times$ speedup and  $\tau = 5.43$.

Furthermore, under zero-shot evaluation using only 40,000 training samples with no benchmark-specific data, DREAM-S continues to outperform both EAGLE-3 and DREAM with an average speedup of 1.72× versus 1.65× and 1.59× respectively, demonstrating that DREAM-S's gains stem from architectural advantages rather than domain-specific data augmentation (Appendix Table~\ref{tab:zero_shot}).
\paragraph{Comparison with Vispec}
We additionally compare DREAM-S against ViSpec~\citep{kang2025vispec},  a concurrent VLM-specific speculative decoding method that accelerates  VLM inference using a lightweight vision adaptor for image token compression  and global visual feature injection into text tokens. To ensure a fair  comparison, we use identical training data (LLaVA-mix665k with 55,000  samples plus 6,000 benchmark-disjoint samples) rather than ViSpec's  synthetic long-response data. As shown in Table~\ref{tab:vispec_main},  DREAM-S outperforms ViSpec across all benchmarks on both  LLaVA-v1.6-Vicuna-7B and 13B, with average speedup improvements of   12.7\% on LLaVA-7B ($2.58\times$ vs.\ $2.29\times$) and 13.8\% on  LLaVA-13B ($3.62\times$ vs.\ $3.18\times$), demonstrating DREAM-S's superior performance under controlled training conditions.
\begin{table}[t]
  \centering
  \small
  \resizebox{\linewidth}{!}{
  \begin{tabular}{llcccccccc}
    \toprule
    \multirow{2}{*}{Model} & \multirow{2}{*}{Method} &
    \multicolumn{2}{c}{MMT} &
    \multicolumn{2}{c}{SEED} &
    \multicolumn{2}{c}{ScienceQA} &
    \multicolumn{2}{c}{Average} \\
    \cmidrule(lr){3-4}\cmidrule(lr){5-6}\cmidrule(lr){7-8}\cmidrule(lr){9-10}
    & & $S$ & $\tau$ & $S$ & $\tau$ & $S$ & $\tau$ & $S$ & $\tau$ \\
    \midrule
    \multirow{2}{*}{\makecell{LLaVA-v1.6 \\ Vicuna-7B}}
    & ViSpec & 2.32 & 3.28 & 2.33 & 3.17 & 2.21 & 2.68 & 2.29 & 3.04 \\
    & \textbf{DREAM-S}   & \textbf{2.67} & 6.27 & \textbf{2.61} & 6.18 & \textbf{2.45} & 5.71 & \textbf{2.58} & 6.05 \\
    \midrule
    \multirow{2}{*}{\makecell{LLaVA-v1.6 \\ Vicuna-13B}}
    & ViSpec & 3.21 & 4.65 & 3.20 & 4.51 & 3.13 & 3.35 & 3.18 & 4.17 \\
    & \textbf{DREAM-S}   & \textbf{3.85} & 5.56 & \textbf{3.61} & 5.32 & \textbf{3.41} & 5.19 & \textbf{3.62} & 5.36 \\
    \bottomrule
  \end{tabular}
  }
  \caption{Evaluation of ViSpec vs.\ DREAM-S on speedup ratio ($S$) and average accepted token length ($\tau$) at temperature $=0$.}
  \label{tab:vispec_main}
\end{table}

\subsection{Ablation Studies}
\begin{figure*}[!t]
   \begin{center}
     \includegraphics[width=1\linewidth]{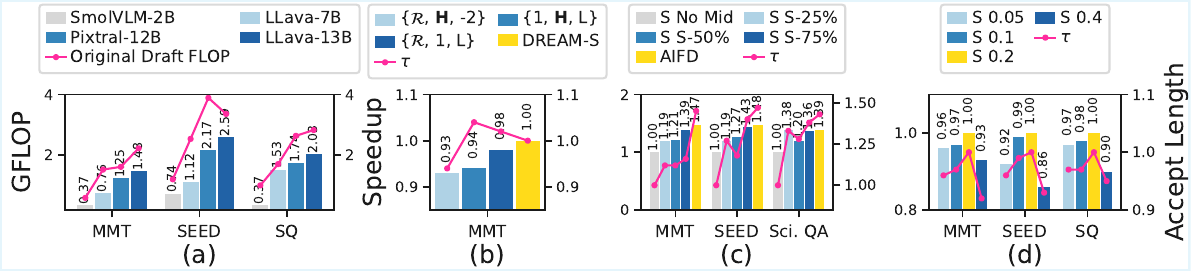}
   \end{center}
   \vspace{-10pt}
      \caption{(a) FLOPs of the selected draft models. (b) DREAM-S performance in various NAS settings. (c) Evaluation of AIFD. (d) Impact of $\lambda$ on DREAM-S performance. In (b)–(d), bars show speedup (left axis) and red curve shows acceptance length (right axis).
}
   \label{fig:ablation}
\end{figure*}
\paragraph{Impact of NAS searching dimension.} To evaluate the impact of each NAS search dimension, we conduct ablation studies on LLaVA-v1.6-Vicuna-7B using MMT-Bench. Specifically, we assess the contributions of searching visual token pruning ($r$), attention head pruning ($H'_{j}$), and adaptive target feature injection ($l'$) by disabling each dimension individually. The results are shown in Figure~\ref{fig:ablation}(b), where $\{\mathcal{R},\textbf{H},-2\}$ denotes searching only for the optimal visual token and head pruning while fixing the draft injection to the second-to-top layer. Similarly, $\{1,\textbf{H},L\}$ and $\{\mathcal{R},1,L\}$ indicate disabling visual token and head pruning search. Figure~\ref{fig:ablation}(b) shows that DREAM-S achieves the highest speedup ($2.67\times$) while maintaining competitive token acceptance ($\tau$ = 6.27). Removing head pruning dimension reduces speedup to $2.62\times$ but increases acceptance length to $\tau$ = 6.44, as the larger draft model with full attention heads captures more information but incurs higher computational cost. Removing visual pruning dimension further decreases speedup to $2.51\times$ while achieving the highest acceptance length ($\tau$ = 6.50), since processing all visual tokens provides complete visual context at the expense of increased latency. Using the target feature injection at fixed position degrades performance ($2.48\times$ speedup), showing that adaptive target feature injection is crucial.



\paragraph{Evaluation of AIFD.} We evaluate the impact of the adaptive intermediate feature distillation (AIFD) strategy described in Section~\ref{sec:intermedaite-distill} on DREAM-S performance. Experiments are conducted using the LLaVA-v1.6-Vicuna-7B model across MMT-Bench, SEED-Bench, and ScienceQA datasets. To show its advantage, we design four baselines: (1)~\textit{No Mid Tuning} (No Mid), which trains DREAM-S without intermediate features; (2)~\textit{Static-25\%} (S-25\%), using features from position of target model's 25\% depth; (3)~\textit{Static-50\%} (S-50\%) extracting from 50\% depth; and (4)~\textit{Static-75\%} (S-75\%) utilizing 75\% depth features.

Figure~\ref{fig:ablation}(c) presents the results. Training without intermediate supervision yields the lowest performance. While static approaches show progressive improvement with depth (Static-75\% strongest), AIFD consistently outperforms all static baselines, confirming that adaptive layer selection based on attention entropy effectively identifies the most informative supervision signals.

\paragraph{Adaptivity of DREAM-S across Different Hardware Conditions.} 
Unlike existing SD baselines that are agnostic to hardware conditions, DREAM-S adapts to different hardware platforms by searching the subnetwork space and adjusting draft model configurations, achieving superior performance across diverse hardware conditions. 
To show this, we evaluate DREAM-S across three different GPU architectures: Nvidia RTX8000 48GB, Nvidia A100 80GB, and Nvidia H100 80GB, representing different hardware conditions. The experiments are conducted on the LLaVA-v1.6-Vicuna-7B model across the MMT-Bench, SEED-Bench, and ScienceQA datasets.
Table~\ref{tab:gpu_latency} indicates that DREAM-S consistently outperforms EAGLE-3 across all hardware configurations. On high-performance GPUs like H100, DREAM-S achieves $2.99\times$ speedup compared to EAGLE-3's $2.67\times$, with throughput reaching 153.12 tokens/s versus 142.25 tokens/s. On resource-constrained hardware like RTX8000, DREAM-S achieves $2.23\times$ speedup while EAGLE-3 drops to $1.88\times$, showing DREAM-S's robustness across diverse hardware environments. 
\begin{table}[!t]
  \resizebox{\linewidth}{!}{%
    \begin{tabular}{c|cc|cc}
      \hline
      \multirow{2}{*}{GPUs} & \multicolumn{2}{c|}{Eagle-3} & \multicolumn{2}{c}{DREAM-S} \\
      & Speedup & Tokens/S & Speedup & Tokens/S \\ \hline 
      A100  & $2.26\times$ & 84.67  & $2.58\times$ & 94.43 \\
      H100  & $2.67\times$ & 142.25 & $2.99\times$ & 153.12 \\ 
      RTX8000  & $1.88\times$ & 37.44  & $2.23\times$ & 43.73\\
      \hline
    \end{tabular}
  }
  \caption{Eagle-3 vs. DREAM-S Performance. }
  \label{tab:gpu_latency}
\end{table}
\paragraph{Impact of Lambda Setting.} As described in equation~\ref{eq:loss}, the loss weights $\lambda_{feat}$, $\lambda_{distill}$, and $\lambda_{KL}$  control the relative importance of the loss function. Since $\mathcal{L}_{feat}$ and $\mathcal{L}_{distill}$ are both smooth L1 losses operating at similar scales with comparable roles in feature alignment, we set $\lambda_{feat} = \lambda_{distill}$ to simplify hyperparameter tuning. The KL divergence loss maintains consistent influence with $\lambda_{KL} = 1$.
Figure~\ref{fig:ablation}(d) demonstrates the impact of varying $\lambda_{feat}$ on LLaVA-v1.6-Vicuna-7B performance across three benchmarks. Each number represents the static (S) value for both $\lambda_{feat}$ and $\lambda_{distill}$, while $\lambda_{KL}$ is fixed to 1. Increasing $\lambda_{feat}$ from 0.05 to 0.2 improves performance metrics across all datasets, indicating that stronger feature supervision enhances draft quality. However, further increasing to 0.4 leads to performance drop, suggesting that excessive feature supervision can impair the model's ability to generalize effectively. This validates our choice of $\lambda_{feat} = \lambda_{distill} = 0.2$, as the optimal balance point.

\section{Conclusion}
In this paper, we introduce \textit{DREAM-S}, a SD framework optimized for VLMs that generalizes robustly across diverse hardware platforms and model scales. By combining neural architecture search and attention-guided feature distillation, DREAM-S achieves up to $3.85\times$ speedup over the existing SD baseline while preserving task performance across diverse multimodal benchmarks.

\section*{Limitations}
Our contribution is primarily system-level and application-driven. Rather than proposing a new generic NAS algorithm or theoretical decoding principle, we instantiate a multimodal, hardware-aware search space for draft design and adopt exhaustive evaluation given its current moderate size. As the space scales to richer architectures or devices, DREAM-S can naturally incorporate more advanced strategies such as OFA-style predictor-based search. Additionally, for architectures with fundamentally different structural components such as MoE-based VLMs, the search dimensions themselves may need to be revisited, as attention head pruning may interact non-trivially with expert routing mechanisms. Extending DREAM-S to such architectures is an interesting future direction.
\color{black}{}


\bibliography{custom}
\newpage
\appendix

\section{Appendix}
\label{sec:appendix}

\paragraph{Two-Phase Progressive Training Algorithm.}
The complete algorithm for our Two-Phase Progressive Training (TPPT) framework is presented in Algorithm~\ref{alg:nas_training}. This algorithm encompasses both the warm-up training phase (Phase 1) and the multi-resolution training phase (Phase 2) described in Section ~\ref{sec:nas_architect}.

\begin{algorithm}[h]
\caption{Two-Phase Progressive Training for drafting (TPPT)}\label{alg:nas_training}
\footnotesize
\begin{algorithmic}
\REQUIRE Training dataset $\mathcal{D}$, supernet draft model $\mathcal{M}_{da}$, target model $\mathcal{M}_{ta}$
\OUTPUT Trained model with optimal subnet architectures
\STATE Initialize supernet parameters $\theta$

\STATE \textbf{Phase 1: Supernet Training}
\FOR{each supernet training epoch}
    \FOR{each batch $\mathcal{B} \in \mathcal{D}$}
        \STATE $T, S \gets \mathcal{M}_{ta}(\mathcal{B})$ \COMMENT{Target model forward pass}
        \STATE $S^{L} \gets S$ \COMMENT{Target's final hidden features}
        \STATE $S^{\ell^\star} \gets \text{AIFD}(S)$ \COMMENT{Adaptive Intermediate Layer distillation}
        \STATE $D, E^{m}, E^{M} \gets \mathcal{M}_{da}(\mathcal{B}, S^{L})$ \COMMENT{Draft model forward pass}
        \STATE Compute Loss($S^{\ell^\star}, E^{m}, S^{L}, E^{M}, T, D$)
        \STATE Update $\theta$ 
    \ENDFOR
\ENDFOR

\STATE \textbf{Phase 2: Subnet Training}
\FOR{each subnet training epoch}
    \FOR{each batch $\mathcal{B} \in \mathcal{D}$}
        \STATE $T, S \gets \mathcal{M}_{ta}(\mathcal{B})$
        \STATE $S^{\ell^\star} \gets AIFD(S)$
        \STATE Sample $r,H'_j,l'$ From $\mathcal{R}, \textbf{H}, L$
        \STATE $\mathcal{B}' \gets \text{Prune}(\mathcal{B}, r)$ \COMMENT{Apply Visual Token Compression}
        \STATE $S^{l'} \gets S$  \COMMENT{Extract Target's feature from selected layer}
        \STATE $D, E^{m}, E^{M} \gets \mathcal{M}_{da}(\mathcal{B'}, H'_j,  S^{l'})$ \COMMENT{Draft Model forward}
         \STATE Compute Loss($S^{\ell^\star}, E^{m}, S^{L}, E^{M}, T, D$)
        \STATE Update $\theta$ 
    \ENDFOR
\ENDFOR
\end{algorithmic}
\end{algorithm}

\begin{table*}[!t]
\centering\resizebox{1.95\columnwidth}{!}{
\begin{tabular}{l|cc|cc|cc|cc|cc|cc}
\toprule
\multirow{2}{*}{Configuration} & \multicolumn{2}{c|}{MMT-Bench} & \multicolumn{2}{c|}{SEED-Bench-2} & \multicolumn{2}{c|}{ScienceQA} & \multicolumn{2}{c|}{ OCRBench} & \multicolumn{2}{c|}{ ChartQA} & \multicolumn{2}{c}{ MathVista} \\
& S & $\tau$ & S & $\tau$ & S & $\tau$ & S & $\tau$ & S & $\tau$ & S & $\tau$ \\
\midrule
Head Pruning Ratio & 2.65 & 6.32 & 2.54 & 6.20 & 2.44 & 6.10 & 2.02 & 4.84 & 1.92 & 4.35 & 2.03 & 5.22 \\
Double Head Pruning Ratio & 2.66 & 6.34 & 2.53 & 6.18 & 2.44 & 6.09 &  2.01 & 4.82 & 1.90 & 4.38 & 2.03 &  5.19 \\
Vision Compression Ratio & 2.67 & 6.40 & 2.60 & 6.22 & 2.57 & 6.02 &  2.07 & 4.88 & 1.94 & 4.44 & 2.13 &  5.32 \\
Double Vision Compression Ratio & 2.66 & 6.37 & 2.62 & 6.22 & 2.57 & 6.00 & 2.09 & 4.85 & 1.95 & 4.43 & 2.12 & 5.28 \\
\bottomrule
\end{tabular}
}
\caption{Analysis of search space granularity on LLaVA-v1.6-Vicuna-7B at Temperature=0}
\label{tab:search_granularity}
\end{table*}

\paragraph{Search Space Granularity Analysis.}
To validate the sufficiency of our chosen search space granularity, we conduct experiments comparing standard and doubled search resolution configurations on LLaVA-v1.6-Vicuna-7B across three benchmarks (Table~\ref{tab:search_granularity}). During the TPPT training phase, both standard and doubled granularity configurations utilize their respective complete search spaces. Our standard Head Pruning Ratio trains with search space $\{0.25, 0.5, 0.75, 1.0\}$, while the Double Head Pruning Ratio configuration trains with the finer grained space $\{0.125, 0.25, 0.375, 0.5, 0.625, 0.75, 0.875, 1.0\}$. Similarly, our standard Vision Compression Ratio trains with the budget pool $\mathcal{R} = \{0.1n, 0.2n, \ldots, n\}$, whereas the Double Vision Compression Ratio trains with $\{0.05n, 0.1n, 0.15n, \ldots, n\}$. After training, we perform exhaustive NAS search across all trained subnetworks to identify the optimal configuration for each benchmark. MMT Bench uses 8,000 training samples, SEED Bench uses 8,000 samples, ScienceQA uses 6,000 samples, OCRBench  uses 8,000 samples, ChartQA uses 8,000 samples and MathVista uses 5,000 samples, with all benchmarks evaluated on 1,000 test samples.

 Table~\ref{tab:search_granularity} demonstrates that doubling the search space granularity yields negligible performance differences, with maximum speedup variation of only 0.02$\times$, confirming that our chosen search space provides adequate coverage of the optimization landscape without requiring computationally expensive fine-grained search. Furthermore, comparing these results with the preliminary experiments in Table~\ref{tab:compression-ablation} reveals DREAM-S's training effectiveness: after full TPPT training, DREAM-S achieves substantially higher speedups (2.65$\times$ vs 2.57$\times$ for head pruning configurations, 2.67$\times$ vs 2.56$\times$ for visual compression) despite maintaining comparable token acceptance lengths. This improvement demonstrates DREAM-S's ability to apply more aggressive pruning strategies during training while preserving draft model quality, enabling superior speed-accuracy trade-offs compared to naive pruning approaches applied without integrated training optimization.

\paragraph{Draft Window Size Analysis.}
Table~\ref{tab:draft_window} reveals the fundamental trade-off in speculative decoding between draft sequence length and computational efficiency. The draft window size determines the maximum number of tokens the draft model can generate before verification by the target model. As $\gamma$ increases from 4 to 8, the average accepted token length ($\tau$) consistently improves across all benchmarks, with MMT-Bench showing an increase from 4.89 to 7.05 tokens. However, this improvement comes at the cost of reduced speedup ratios, which decline from 2.76$\times$ to 2.59$\times$ on MMT-Bench. This trade-off occurs because when draft sequences are rejected, the computational cost of generating all the rejected tokens is wasted, and rejection typically happens early in the sequence, rendering most subsequent tokens in longer draft windows wasteful. The diminishing speedup returns beyond $\gamma$=6 suggest an optimal balance point where the computational overhead of generating additional draft tokens begins to outweigh the benefits of potentially longer accepted sequences. This analysis validates our choice of $\gamma$=6 in the main experiments and demonstrates that effective speculative decoding requires careful calibration between speculation aggressiveness and computational efficiency.

\begin{table}[h]
\centering\resizebox{1\columnwidth}{!}{
\begin{tabular}{c|cc|cc|cc}
\toprule
\multirow{2}{*}{$\gamma$ (Draft Window Size)} & \multicolumn{2}{c|}{MMT-Bench} & \multicolumn{2}{c|}{SEED-Bench-2} & \multicolumn{2}{c}{ScienceQA} \\
& S & $\tau$ & S & $\tau$ & S & $\tau$ \\
\midrule
4 & 2.76 & 4.89 & 2.68 & 4.79 & 2.53 & 4.87 \\
5 & 2.73 & 5.78 & 2.69 & 5.62 & 2.56 & 5.09 \\
6 & 2.67 & 6.27 & 2.61 & 6.18 & 2.45 & 5.71 \\
7 & 2.62 & 6.72 & 2.55 & 6.54 & 2.41 & 5.93 \\
8 & 2.59 & 7.05 & 2.51 & 7.07 & 2.36 & 6.37 \\
\bottomrule
\end{tabular}
}
\caption{Evaluation of draft window size ($\gamma$) impact on DREAM-S performance for LLaVA-v1.6-Vicuna-7B at Temperature=0.}
\label{tab:draft_window}
\end{table}

\begin{table*}[!t]
\centering\resizebox{1.95\columnwidth}{!}{
\begin{tabular}{l|cc|cc|cc|cc|cc|cc|cc}
\toprule
\multirow{2}{*}{Method} & \multicolumn{2}{c|}{MMT-Bench} & \multicolumn{2}{c|}{SEED-Bench-2} & \multicolumn{2}{c|}{ScienceQA} & \multicolumn{2}{c|}{OCRBench} & \multicolumn{2}{c|}{ChartQA} & \multicolumn{2}{c|}{MathVista} & \multicolumn{2}{c}{Average} \\
& S & $\tau$ & S & $\tau$ & S & $\tau$ & S & $\tau$ & S & $\tau$ & S & $\tau$ & S & $\tau$ \\
\midrule
EAGLE-3 & 1.87 & 4.55 & 1.86 & 4.48 & 1.70 & 4.18 & 1.41 & 3.38 & 1.40 & 3.41 & 1.28 & 3.08 & 1.59 & 3.85 \\
DREAM   & 1.92 & 4.66 & 1.90 & 4.63 & 1.74 & 4.24 & 1.47 & 3.50 & 1.56 & 3.78 & 1.30 & 3.12 & 1.65 & 3.99 \\
\rowcolor{blue!10}
\textbf{DREAM-S} & \textbf{2.01} & 4.61 & \textbf{1.99} & 4.59 & \textbf{1.81} & 4.20 & \textbf{1.51} & 3.47 & \textbf{1.65} & 3.77 & \textbf{1.35} & 3.13 & \textbf{1.72} & 3.96 \\
\bottomrule
\end{tabular}
}
\caption{ Zero-Shot Performance Comparison on LLaVA-Vicuna-7B (Temperature = 0) on speedup ratio (S) and average accepted token length ($\tau$).}
\label{tab:zero_shot}
\end{table*}

\begin{figure*}[!t]
   \begin{center}
     \includegraphics[width=0.7\textwidth]{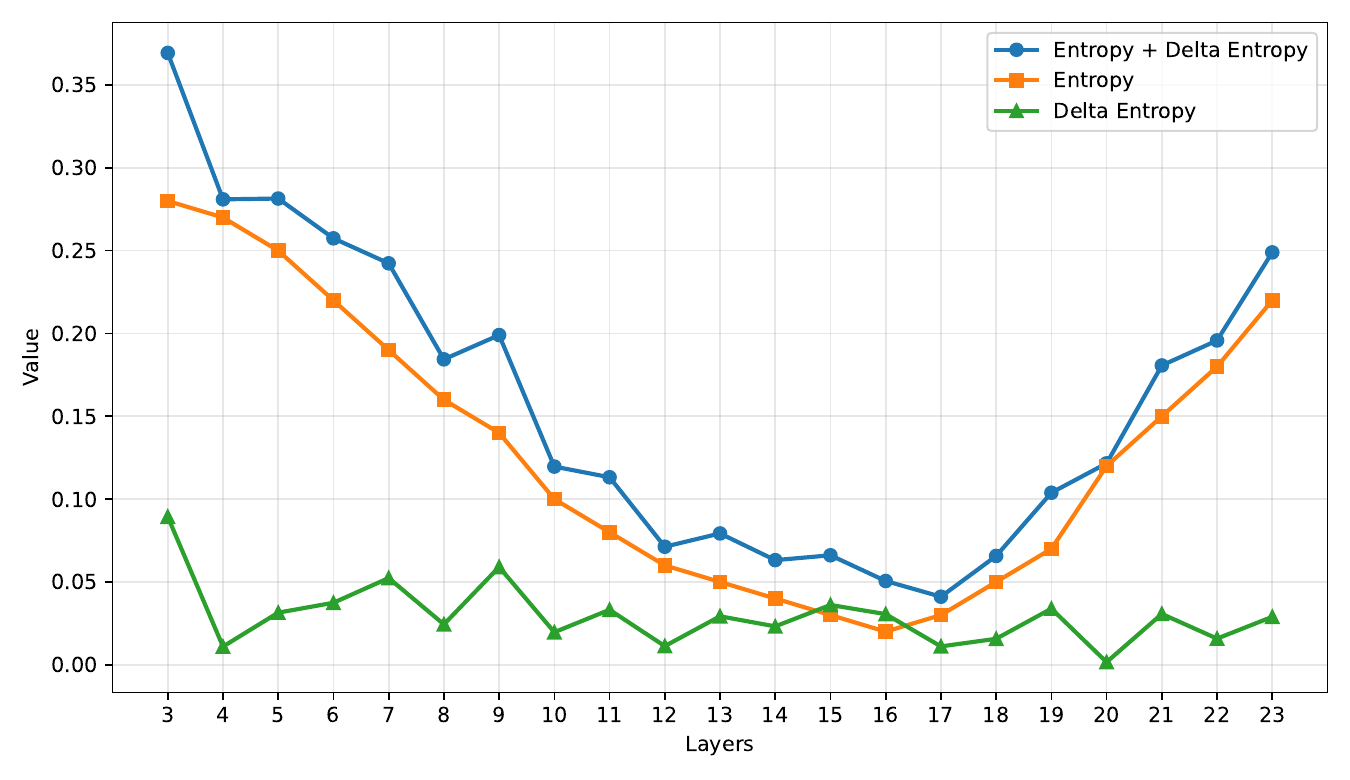}  
   \end{center}
   \caption{Layer-wise comparison of entropy, $\Delta$entropy, and their sum.}
   \label{fig:entropy}
\end{figure*}

\begin{table*}[!t]
    \vspace{0pt}
    \centering\resizebox{1.8\columnwidth}{!}{
    {\huge
       \begin{tabular}{cccccccccccccc|cc}
        \toprule
        &  & \multicolumn{2}{c}{MMT} & \multicolumn{2}{c}{SEED} & \multicolumn{2}{c}{ScienceQA} & \multicolumn{2}{c}{OCRBench} & \multicolumn{2}{c}{ChartQA} & \multicolumn{2}{c}{MathVista}  & \multicolumn{2}{|c}{Average} \\   \midrule
    Models & Methods & S & $\tau$ & S & $\tau$ & S & $\tau$ & S & $\tau$ & S & $\tau$ & S & $\tau$ & S & $\tau$   \\ \midrule
    \multirow{2}{*}{LLaVA-Vicuna-7B}
        & DREAM-S (Fixed) & 2.67 & 6.27 & 2.54 & 6.03 & 2.39 & 5.57 & 2.07 & 4.78 & 1.94 & 4.31 & 2.13 & 5.23 & 2.29 & 5.39 \\
        & \textbf{DREAM-S} & \textbf{2.67} & 6.27 & \textbf{2.61} & 6.18 & \textbf{2.45} & 5.71 & \textbf{2.11} & 4.89 & \textbf{2.04} & 4.39 & \textbf{2.20} & 5.30 & \textbf{2.35} & 5.46 \\
    \midrule
    \multirow{2}{*}{LLaVA-Vicuna-13B}
        & DREAM-S (Fixed) & 3.85 & 5.56 & 3.56 & 5.21 & 3.40 & 5.24 & 2.70 & 4.54 & 2.61 & 4.07 & 2.56 & 4.06 & 3.11 & 4.78 \\
        & \textbf{DREAM-S} & \textbf{3.85} & 5.56 & \textbf{3.61} & 5.32 & \textbf{3.41} & 5.19 & \textbf{2.77} & 4.61 & \textbf{2.67} & 4.17 & \textbf{2.62} & 4.11 & \textbf{3.16} & 4.83 \\
    \midrule
    \multirow{2}{*}{Pixtral-12B}
        & DREAM-S (Fixed) & 2.99 & 4.48 & 2.71 & 3.48 & 3.04 & 3.84 & 2.39 & 3.36 & 2.37 & 3.35 & 2.38 & 3.23 & 2.65 & 3.62 \\
        & \textbf{DREAM-S} & \textbf{3.01} & 4.41 & \textbf{2.73} & 3.56 & \textbf{3.09} & 3.93 & \textbf{2.46} & 3.44 & \textbf{2.40} & 3.42 & \textbf{2.42} & 3.34 & \textbf{2.69} & 3.68 \\
    \midrule
    \multirow{2}{*}{SmolVLM-2B}
        & DREAM-S (Fixed) & 3.12 & 3.94 & 2.26 & 3.14 & 2.88 & 3.54 & 1.88 & 2.51 & 1.63 & 2.26 & 2.04 & 2.79 & 2.30 & 3.03 \\
        & \textbf{DREAM-S} & \textbf{3.12} & 3.94 & \textbf{2.28} & 3.16 & \textbf{2.91} & 3.57 & \textbf{1.88} & 2.51 & \textbf{1.64} & 2.28 & \textbf{2.06} & 2.82 & \textbf{2.32} & 3.05 \\
    \bottomrule
    \end{tabular}}
}
\vspace{2pt}
\caption{Evaluation of DREAM-S with Fixed Configuration compared with DREAM-S on speedup ratio (S) and average accepted token length ($\tau$).}
\label{tab:fixed_config}
\end{table*}
\paragraph{Entropy and Delta Entropy Visualization.}
As shown in Figure~\ref{fig:entropy}, while entropy captures the average uncertainty of token-level attention distributions, it reflects how dispersed or concentrated the attention is in each layer. However, model stability and information flow depend on how that entropy changes across layers. $\Delta$ entropy highlights fluctuations, revealing whether a layer’s attention is becoming more stable or more chaotic relative to its neighbors.
If we used only entropy, we would overlook these dynamic shifts and miss layers where abrupt structural changes or information transitions occur. By summing entropy and $\Delta$ entropy, the total metric integrates both the static view of uncertainty and the dynamic view of its variation, providing a more faithful signal for selecting layers and guiding downstream decisions.

\paragraph{Zero-Shot Generalization.}
To assess DREAM-S's generalization beyond domain-adapted training, we retrain DREAM-S, DREAM, and EAGLE-3 using only 40K samples from LLaVA-Mix665K, excluding all evaluation benchmark samples. As shown in Table~\ref{tab:zero_shot}, performance drops modestly across all three methods due to reduced training data, confirming this is a shared property of speculative decoding approaches rather than a limitation unique to DREAM-S. Even under this stricter zero-shot setting, DREAM-S continues to achieve the highest average speedup of $1.72\times$, outperforming DREAM ($1.65\times$) and EAGLE-3 ($1.59\times$). This demonstrates that DREAM-S's gains stem from its architectural advantages, including NAS-driven draft optimization, adaptive visual token compression, and attention-entropy-guided distillation, rather than from domain-specific data augmentation.

\paragraph{Generalizability with Fixed Configuration.}
To evaluate whether DREAM-S's gains depend on per-dataset NAS tuning, we test a single fixed draft configuration applied uniformly across all four VLMs and six benchmarks, with no dataset-specific adaptation. Table~\ref{tab:fixed_config} reports speedup ratios (S) and average accepted token lengths ($\tau$) for this fixed-configuration variant (DREAM-S Fixed) alongside the dataset-adaptive DREAM-S.

The fixed configuration still decisively outperforms all baselines, including the strongest competitors EAGLE-3 and DREAM (see Table~\ref{tab:eval:main}). For example, on LLaVA-13B, DREAM-S Fixed achieves $3.11\times$ average speedup versus EAGLE-3's $2.89\times$ and DREAM's $3.06\times$. The performance gap between fixed and adaptive configurations is most noticeable on fine-grained visual benchmarks such as OCRBench, ChartQA, and MathVista, where aggressive visual token pruning can discard detail needed for character- or chart-level precision.  These results confirm that DREAM-S's improvements stem from genuine architectural and training advances rather than just from dataset-specific configuration tuning, while also demonstrating that the NAS search adds meaningful value for tasks with heterogeneous visual complexity.

\end{document}